\pdfoutput=1
\documentclass[11pt]{article}
\usepackage[hyperref]{acl}
\usepackage{times}
\usepackage{enumitem}
\usepackage{latexsym}
\usepackage[T1]{fontenc}
\usepackage[utf8]{inputenc}
\usepackage{microtype}
\usepackage{tabularx}
\usepackage{booktabs}
\usepackage{mparhack}
\newcommand{\eg}{\textit{e.g.}\xspace}
\usepackage{bbm}
\usepackage{amsmath}

\usepackage{multirow}
\usepackage{scalefnt}
\usepackage{graphicx}
\newcommand{\rt}[1]{\rotatebox{90}{#1}}
\usepackage{xspace}
\newcommand{\F}{F$_1$\xspace}
\definecolor{rosegold}{rgb}{0.72, 0.43, 0.47}
\definecolor{viridian}{rgb}{0.25, 0.51, 0.43}
\definecolor{stildegrainyellow}{rgb}{0.98, 0.85, 0.37}
\renewcommand{\paragraph}[1]{\noindent\textbf{#1}}
\usepackage{tikz}
\usetikzlibrary{shapes.geometric, arrows}

\title{Items from Psychometric Tests as\\ Training Data for Personality
  Profiling Models of Twitter Users}

\author{Anne Kreuter$^1$, Kai Sassenberg$^{2,3}$, \and Roman Klinger$^1$ \\
  $^1$Institut f{\"u}r Maschinelle Sprachverarbeitung, University of Stuttgart, Germany \\
  $^2$Leibniz-Institut f\"ur Wissensmedien, T\"ubingen, Germany\\
  $^3$University of T\"ubingen, Germany\\
  \texttt{\{anne.kreuter,roman.klinger\}@ims.uni-stuttgart.de}\\
  \texttt{k.sassenberg@iwm-tuebingen.de}\\
}

\begin{document}
\maketitle
\begin{abstract}
  Machine-learned models for author profiling in social media often
  rely on data acquired via self-reporting-based psychometric tests
  (questionnaires) filled out by social media users. This is an
  expensive but accurate data collection strategy. Another, less
  costly alternative, which leads to potentially more noisy and biased
  data, is to rely on labels inferred from publicly available
  information in the profiles of the users, for instance self-reported
  diagnoses or test results.
  In this paper, we explore a third strategy, namely to directly use a
  corpus of items from validated psychometric tests as training
  data. Items from psychometric tests often consist of sentences from
  an I-perspective (\eg, ``I make friends easily.''). Such corpora of
  test items constitute `small data', but their availability for many
  concepts is a rich resource.
  We investigate this approach for personality profiling, and evaluate
  BERT classifiers fine-tuned on such psychometric test items for the
  big five personality traits (openness, conscientiousness,
  extraversion, agreeableness, neuroticism) and analyze various
  augmentation strategies regarding their potential to address the
  challenges coming with such a small corpus.
  Our evaluation on a publicly available Twitter corpus shows a
  comparable performance to in-domain
  training for 4/5 personality traits with T5-based data
  augmentation.
\end{abstract}

\section{Introduction}
The field of author profiling originally emerged from the study of
stylometry \cite{lutoslawski1898principes} and, with the rise of
social media \cite{bilan2016caps}, now considers a variety of
attributes, including demographic data such as age, sex, gender,
nationality \cite{schwartz2013personality}, personality traits
\cite{golbeck2011predicting}, or psychological states such as
emotions, or medical conditions like mental disorders
\cite{de2013predicting}. Such automatic methods enable large-scale
social media data analyses even for (combinations of) variables for
which results from surveys are not available. Therefore, personality
profiling in social media helps to paint a more comprehensive,
complete, and timely picture for parts of a society.

State-of-the-art models reconstruct personality traits or mental
health states from posts of social media users by relying on
ground-truth data that links such posts to the correct annotation
\cite{Guntuku2017}. The ground-truth data is typically obtained by
either (1) asking participants to complete a validated survey that
measures the desired variable and asking the participants to share
their social media profiles, (2), by relying on self-reports of users,
e.g., disclosure of a condition in the user's profile description, or
(3), by having experts annotate profiles for particular
properties. The quality of data obtained might therefore suffer from
social-desirability bias, from being a non-representative subsample,
from a lack of validated diagnoses, or from noise stemming from the
challenge that annotators do not have access to the actual
characteristics of users \cite{ernala2019methodological}.

\begin{table}[b]
  \centering\small\scalefont{1.1}
  \setlength{\tabcolsep}{3.9pt}
  \begin{tabularx}{\linewidth}{lcX}
    \toprule
    Variable & Cor. & Item Text \\
    \cmidrule(r){1-1}\cmidrule(rl){2-2}\cmidrule(l){3-3}
    Openness     & $+$ & ``Am interested in many things.''\\
    Openness     & $-$ & ``Do not like art.''\\
    Extraversion & $+$ & ``Warm up quickly to others.''\\    
    Extraversion & $-$ & ``Am hard to get to know.''\\    
    \bottomrule
  \end{tabularx}
  \caption{Example items from a psychometric test to assess
    personality traits \citep{lee2018psychometric}. `Cor.'
    indicates if the item has been shown to correlate positively or
    negatively to the respective concept.}
  \label{tab:psychoexamples}
\end{table}

We explore another route for which we hypothesize that it addresses
these issues, but at the cost of only having access to very small data
sets: We propose to leverage the existing set of high-quality,
validated, and reliable psychometric instruments to measure
psychological traits \textit{directly}. Psychometric tests often come
in the form of questionnaires which contain items, allowing a person
to report about themselves. These items are sentences formulated as
descriptions of the self (Table \ref{tab:psychoexamples} shows some
examples). This structure motivates our hypothesis that such
psychometric tests can be used directly to induce classifiers that
profile individuals in social media without the existence of
designated, manually annotated in-domain training data. If indeed
possible, this would lead to a straight-forward route to develop a
myriad of classifiers for all those concepts for which psychometric
tests exist.  To dampen the issue of these sets of items being
comparably small, we make use of pre-trained language models
\cite{howard-ruder-2018-universal,devlin-etal-2019-bert,Brown2020} to
transfer knowledge acquired through pretraining rich semantic
representations. Some subtypes of such models can be considered
few-shot learners \citep{Brown2020,Ruder2019}, however, the transfer
might not be successful to data outside of the pretraining
domain. Therefore, we evaluate if various data augmentation methods
can further leverage the challenges coming with such small corpora.

Thus, our contributions in this paper are that we (1) assemble a
corpus from publicly available psychometric tests for the `Big Five'
variables of openness, conscientiousness, extraversion, agreeableness,
and neuroticism \cite{costa1992neo}, which have been shown to be
principled factors of personality \citep{cattell1945description}.
Based on these data, we (2) fine-tune BERT
\cite{devlin-etal-2019-bert} and evaluate it on an existing
personality trait corpus \cite{rangel2015overview}. Furthermore, (3)
we evaluate three data augmentation methods, namely paraphrasing with
T5 \cite{t5_2020}, and item generation with GPT-2
\cite{radford2019language} and synonym replacements with Easy Data
Augmentation \cite{wei-zou-2019-eda}. Our results, (4), show that the
models perform en par with in-domain training for 4/5 personality
trait variables.

\section{Related Work}
\paragraph{Psychometric Personality Tests.} A psychometric test is a
standardized instrument used to measure the cognitive, behavioral, or
emotional characteristics of a person. One possible form are
questionnaires, which can be designed for self-reporting. For each
item the information is available if it is correlated positively or
negatively with the concept to be measured. Publicly available
psychometric tests can be found in various online
repositories.\footnote{\url{https://psychology-tools.com/},
  \url{https://ipip.ori.org/},\\
  \url{https://www.psychometrictest.org.uk/}}

An established test for personality traits following the so-called
`big five' variables is the \textit{International Personality Item
  Pool Representation} of the NEO PI-R with 300
items\footnote{\url{http://www.personal.psu.edu/~j5j/IPIP/}}
\citep[IPIP-NEO-300, ][]{goldberg1999broad}. This test is a proxy of
the \textit{Revised NEO Personality Inventory} (NEO PI-R) by
\citet{costa1992neo}, which is copyrighted and can only be ordered by
professionals and used with permission. We use all items of the
IPIP-NEO-300 as the source of our training corpus.

Another test of personality traits would be the \textit{HEXACO Personality
Inventory-Revised} \cite{lee2008hexaco}. It measures six factors of
personality \citep{ashton2004six} with 200 items, namely
Honesty-Humility, Emotionality, Extraversion, Agreeableness,
Conscientiousness, and Openness to experience.

\paragraph{Data.} Psychometric tests found application in the analysis
of social media user's personality in the past. An influential study
has been the work by \citet{schwartz2014towards},
who collected Facebook data with a dedicated application
\cite{stillwell2004mypersonality} in which users completed the
100-item IPIP-NEO-100 questionnaire \cite{goldberg1999broad}. The
users further shared access to their status updates. This data is not available any longer.

The data for the PAN-author-profiling shared task in 2015 has been
collected in a similar way
\citep{rangel2015overview}.\footnote{\url{https://zenodo.org/record/3745945},\url{https://pan.webis.de/}}
It consists of Tweets of 294 English Twitter profiles (besides
Spanish, Italian and Dutch Twitter profiles), which are annotated with
gender, age, and the `Big Five' personality traits. The personality
traits were self-assessed by the Twitter users with the BFI-10
\cite{rammstedt2007measuring}, which is an economic psychometric test
that allows the personality to be recorded with only 10 items. We use
this corpus for evaluation.

\paragraph{Combining Tests and Social Media Data.} An interesting
combination of psychometric tests with social media posts, which is
likely the one most similar to our paper, is the work by
\citet{vu-etal-2020-predicting}. The authors make use of social media
data of users to automatically fill the IPIP-NEO
\cite{goldberg1999broad} psychometric test to predict the social media
user's `Big Five' personality traits. They do so by embedding
sentences and items with BERT into the same distributional space,
followed by a $k$-nearest-neighbor classification. This approach
constitutes the opposing approach that we chose in our paper --
\citet{vu-etal-2020-predicting} use social media data to fill a
psychometric test. We use psychometric tests to classify social media
data.

We refer the reader to \citet{Stajner2020} for a more comprehensive
overview of related work.

\begin{figure}[t]
  \centering
  \sf\small
  \tikzstyle{model} = [rectangle, rounded corners, minimum width=20mm, minimum height=1cm,text centered, draw=black, fill=red!30]
  \tikzstyle{io} = [trapezium, trapezium left angle=70, trapezium
  right angle=110, minimum width=1mm, minimum height=1cm, text
  centered, text width=1.2cm, draw=black, fill=blue!30]
  \tikzstyle{arrow} = [thick,->,>=stealth]
  \tikzstyle{process} = [rectangle, minimum width=20mm, minimum height=1cm, text centered, text width=2.5cm, draw=black, fill=orange!30]
  \tikzstyle{decision} = [diamond, minimum width=20mm, minimum height=1cm, text centered,text width=1cm, draw=black, fill=green!30]
  \tikzstyle{arrow} = [thick,->,>=stealth]
  
  \resizebox{0.8\linewidth}{!}{%
    \begin{tikzpicture}[node distance=15mm]
      \node (in1) [io] {Original Items};
      \node (pro1) [process, below of=in1] {Data\\ Augmentation};
      \draw [arrow] (in1) -- (pro1);
      \node (out1) [io, below of=pro1] {Augm. Items};
      \draw [arrow] (pro1) -- (out1);
      \node (pro2) [process, below of=out1, xshift=0mm] {Fine-tune BERT};
      \draw [arrow] (out1) -- (pro2);
      \node (model) [model, right of = pro2, xshift=20mm, yshift=10mm] {Model};
      \draw [arrow] (pro2) -- (model);
      \node (tweets) [io, above of = model] {Individual\\ Tweets};
      \draw [arrow] (tweets) -- (model);
      \node (in2) [io, above of = tweets] {Twitter Profiles};
      \draw [arrow] (in2) -- (tweets);
      \node (aggreg) [process, below of = model] {Aggregated\\ Evaluation};
      \draw [arrow] (model) -- (aggreg);
      \node (out2) [io, below of = aggreg] {Evaluation\\{\ }};
      \draw [arrow] (aggreg) -- (out2);
      \draw [arrow] (in2) to [out=340,in=30] (aggreg) node[xshift=21mm,yshift=24mm,rotate=270]{gold labels};
    \end{tikzpicture}
  }
  \caption{Workflow of our approach.}
  \label{fig:pipeline}
\end{figure}
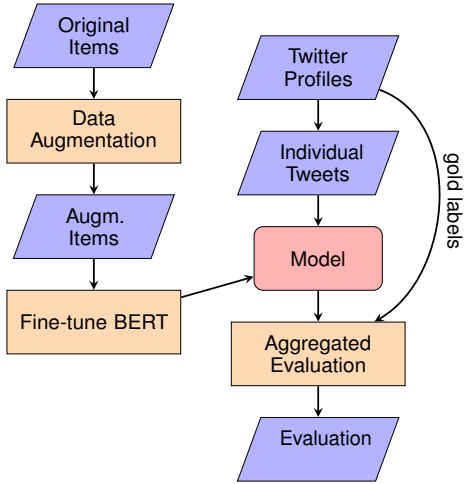

\section{Methods}
\subsection{Workflow}
We depict the general workflow in Figure~\ref{fig:pipeline}. The
original items from the questionnaire are first augmented. The
resulting augmented items inherit the labels from the respective
original items. We fine-tune BERT with these items which leads to a
model to make predictions for comparably short instances, like
Tweets. From the labeled corpus of Twitter profiles, we obtain labels
for each individual tweet with the BERT-based model and then aggregate
the individual labels to obtain a label for the whole profile. In the
evaluation, this predicted profile label is compared to the annotated
gold label.

\subsection{Corpora}
We use all items of the psychometric test IPIP-NEO-300
\citep{goldberg1999broad} as training data and label each item
following the evaluation guidelines accompanying the IPIP-NEO-300 (see
also Table \ref{tab:psychoexamples}). These guidelines provide the
information if a confirmative answer to the item indicates a positive
correlation or negative correlation with the target variable, which
leads to a binary label.

For evaluation, we use the English subset of the
PAN-author-profiling-2015 data \cite{rangel2015overview} with
annotated Twitter profiles. Table \ref{tab:corpusstatistics}
summarizes the corpus statistics. Note that the distribution of the
items from the test data is skewed towards positive instances -- this
might be a direct consequence of people with particular personality
traits being more likely to share particular information on social
media.

\begin{table}
  \centering\small
  \setlength{\tabcolsep}{5.3pt}
  \renewcommand*{\arraystretch}{1.0}
  \begin{tabular}{l rr rr rr}
    \toprule
    & \multicolumn{2}{c}{IPIP-NEO} 
    & \multicolumn{2}{c}{Profiles} 
    & \multicolumn{2}{c}{Tweets}\\
    \cmidrule(lr){2-3}\cmidrule(lr){4-5}\cmidrule(l){6-7}
    Class. & \multicolumn{1}{c}{$+$}
    & \multicolumn{1}{c}{$-$}
    & \multicolumn{1}{c}{$+$}
    & \multicolumn{1}{c}{$-$}
    & \multicolumn{1}{c}{$+$}
    & \multicolumn{1}{c}{$-$}
    \\
    \cmidrule(lr){2-3}\cmidrule(lr){4-5}\cmidrule(l){6-7}
    Open.      &28&32&288&3&26,743&236   \\
    Consc. &31&29&229&15&21,391&1,428   \\
    Extra.  &36&24&235&21&21,686&2,000   \\
    Agree. &24&36&223&29&20,441&2,831   \\
    Neurot.   &33&27&76&197&18,076&7,168   \\
    \bottomrule
  \end{tabular}
  \caption{Corpus Statistics regarding the Twitter evaluation data and
    the IPIP-NEO-300-based training corpus. We extracted all profiles with positive or negative scores and excluded profiles with neutral scores.}
  \label{tab:corpusstatistics}
\end{table}

\subsection{Classification Model}
As our source domain, we consider a set of items
$Q_C = \{(q_i,y_i)\}_{i=1}^n$ from a reliable psychometric test. Each
of these items corresponds to one psychological concept $C$ and
consists of the item text $q_i$ and the label
$y_i \in \{\mathrm{pos}, \mathrm{neg}\}$ which stems from the
evaluation guidelines for this test.

The task is to find a parameterized function $f_{C,\lambda}(u)$ which
takes as input all posts of a user $u$ and predicts a label for each
concept $C$. The important aspect in our setup is that the parameters
$\lambda$ are only optimized on the psychometric data $Q_C$. This is a
mismatch -- we train a classifier to label short texts but need as
output a prediction for a set of tweets which represents the
user. Hence, to obtain a label for each user, we aggregate the labels
for all their posts by accepting the majority class, for each concept
separately.

To obtain the text classifier, we fine-tune BERT
\citep{devlin-etal-2019-bert} to approximate each function
$f_{C,\lambda}$, based on \textit{bert-base-uncased}. The sequence
classification head is randomly initialized on top of the
encoder.\footnote{\url{https://huggingface.co/transformers/model_doc/bert.html\#bertforsequenceclassification}}
For each concept $C$, we fine-tune a separate BERT model (no
multi-task learning).

\begin{figure*}[t]
  \centering
  \includegraphics[scale=0.98]{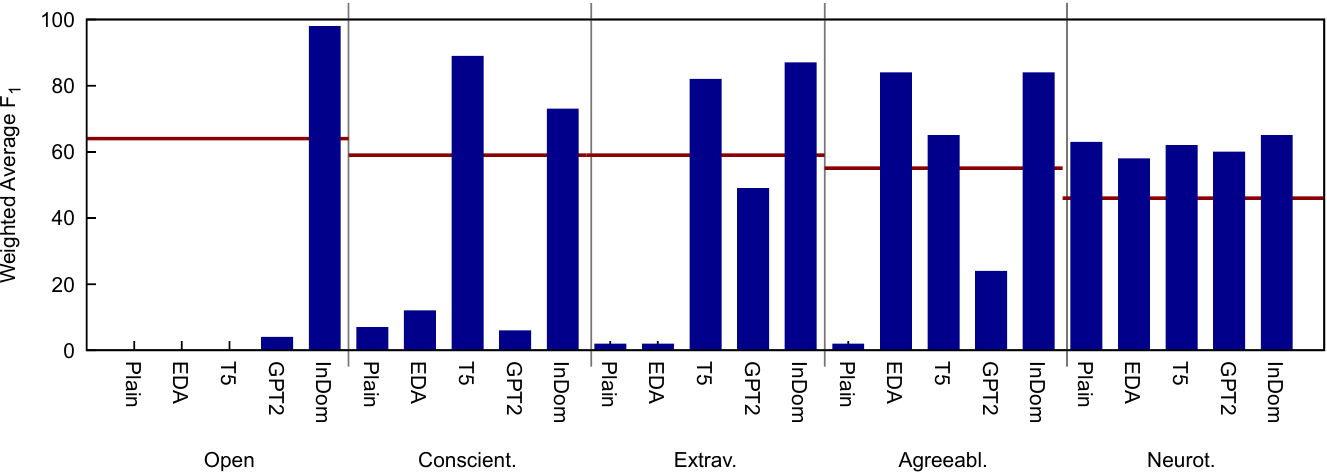}
  \caption{\F scores for all models and classes. Horizontal
    lines depict the baseline.}
  \label{fig:results} 
\end{figure*}

\subsection{Data Augmentation}

With 60 items per personality trait, our
training corpora are small. To address this issue, we perform data
augmentation with three different methods.
For every $n$ instances $(q_i,y_i) \in Q_C$, we perform each data
augmentation $m$ times (obtaining $n\cdot{}m$ instances). Thus, we
generate $m$ augmented items $q^a_i$ for each 
$q_i$. Each newly generated instance inherits the label $y_i$ of its
original instance $q_i$. 
We show examples for automatically generated items in the Appendix
\ref{sec:augmentation}.

\paragraph{Easy Data Augmentation.}
Easy Data Augmentation \cite[EDA,][]{wei-zou-2019-eda} consists of
four operations on the sentence level: synonym replacement, random
insertion, random deletion, and random swap. We use the default
parameter of 10\% of words in the sentence being changed (30\% for
random deletion) to perform each operation of EDA on each sentence
(item) 5 times, hence generate 20 instances out of each original
instance. This leads to 1,160 items for openness, 1,130 for
conscientiousness, 1,080 for extraversion, 1,160 for agreeableness,
and 1080 for neuroticism.

\paragraph{T5 item paraphrasing.}
We use T5 \cite{t5_2020} to paraphrase each item, based on the
\textit{T5ForConditionalGeneration} model provided by
HuggingFace\footnote{\url{https://huggingface.co/transformers/model_doc/t5.html\#t5forconditionalgeneration}}.
We do not perform fine-tuning to our domain, but rely on the original
pre-trained parameters. For each item, we generate up to 50
paraphrases which leads to 2,285 items for openness, 2,383 for
conscientiousness, 2,149 for extraversion, 2,126 for agreeableness,
2,130 for neuroticism.

\paragraph{GPT-2 item generation.}
We fine-tune GPT-2 \cite{radford2019language} for each personality
trait separately in 150 epochs, based on
gpt-2-simple\footnote{\url{https://github.com/minimaxir/gpt-2-simple}}.
We generate 3000 items for each class label with a sentence length of
100 tokens and a temperature of 1.5.  This leads to 6,279 items for
openness, 6,177 for conscientiousness, 6,204 for extraversion, 6,271
for agreeableness, 6,242 for neuroticism.

\section{Experiments}
\subsection{Experimental Settings}
We split the psychometric test data to 80\,\% for training and use
20\,\% for hyperparameter optimization, while we ensure that augmented
items stay in one set with their original item.\footnote{Seed set to
  42 via PyTorchLighting \textit{seed\_everything}, learning rate
  $10^{-5}$, batch size of 16, optimization with Adam
  \cite{kingma2014adam}.}  To avoid overfitting, we apply early
stopping via observing the loss on the validation data. The maximum
number of epochs is set to 200.

For a comparison to an ``upper-bound'' of in-domain training on
Twitter, we split the corpora of social media profiles such that 50\%
of the Twitter profiles are in the test set. The remaining 50\,\% are
used for training and further split into 90\,\% for training and
10\,\% for hyperparameter optimization of the in-domain model. The
settings for fine-tuning the in-domain models are identical to the
settings of the psychometric models.

\subsection{Results}
We show our main results as weighted \F values in
Figure~\ref{fig:results} (complete results in Table~\ref{tab:results}
in the Appendix). We compare the ``plain'' models without data
augmentation to the augmented methods (as bar plots) and a random
baseline (as horizontal line). We further show the performance of the
in-domain model.

All ``plain'', non-data augmented models get outperformed by the
random baseline, except for the personality trait neuroticism (\F=.63
versus \F=.46 random baseline).  The plain psychometric models are
inferior to the in-domain models for all concepts, but to various
extends: Neuroticism is the only trait where the plain model shows a
performance \textit{en par} with the in-domain model.

Regarding the augmentation methods, T5 shows considerable improvements
for conscientiousness, extraversion, and agreeableness (\F=.89,
\F=.82, \F=.65, respectively, vs. .73, .87, .84 for in-domain
models). This is also the best-performing augmentation method for
conscientiousness and extraversion, however, EDA shows a further
improvement for agreeableness (.84). T5 does not harm the
performance for neuroticism in comparison to the plain
model. Therefore, we conclude that T5 augmentation is a promising
choice for 4/5 traits, while the other augmentation methods appear
less stable in their contribution.

In summary, we obtain a substantial model performance without the use
of in-domain training data for Conscientiousness, Extraversion,
Agreeableness, and Neuroticism. The transfer to or the difficulty of
these concepts appears not to be the same, the performance for
Conscientiousness is substantially higher than for Neuroticism. These
results can only be partially compared to previous work due to the
differences in the evaluation setup. However, it should be noted that
the concepts that appear to be more challenging in our setup show
also lower evaluation measures in related work \cite[see for instance
Table 3 in][note that their evaluation measure is an \textsc{rsme},
lower is therefore better]{rangel2015overview}.

\subsection{Model Introspection}
To provide some insights on the decision process by the classification
models, we provide one example for each personality trait from the
Tweet corpus with LIME explanations \citep{lime} in
Table~\ref{tab:lime}. In the example for openness, the classifier
relies on the word ``love'' as a positive indicator. This word can indeed be
found in items from the test, namely in ``Love to daydream'', ``Love
flowers'', and ``Love to read challenging material''. It is also a
term that appears frequently in augmented data, such as in ``Love
problem solving'' or in ``Love flowers. Is it not hard to tell if you
like something that’s especially beautiful?''. A positive indicator
for conscientiousness is the word ``August'' and ``year''. This is
interesting, given that these words appear not to be directly related
to conscientiousness, and they do not appear in the original items of
the test. However, the augmented data contains items that refer to
``year'', such as in ``I truly love Excel and have used it for
years.''.

\begin{table}
  \centering\small\scalefont{1.0}
  \begin{tabularx}{\linewidth}{lX}
    \toprule
    T & Tweet \\
    \cmidrule(r){1-1}\cmidrule(l){2-2}
    O & @\colorbox{viridian!60}{username} What my \colorbox{viridian!30}{love} life will hold instore for me in the future. I'd \colorbox{rosegold!40}{never} ask \colorbox{rosegold!40}{when} I'm gonna die...???????? \\
    C & ``@username: @username I like your profile photo.  \colorbox{rosegold!40}{Very} nice!!! You look \colorbox{rosegold!40}{very} pretty.  :)" THANK YOU! Took this photo in \colorbox{viridian!60}{August} this \colorbox{viridian!30}{year}. \\
    E & @\colorbox{rosegold!40}{username} Slade!!! \colorbox{viridian!60}{Cool} memories of my \colorbox{rosegold!40}{grammar} \colorbox{viridian!30}{school} days!! \\
    A & @username I rocked so much \colorbox{viridian!60}{to} \colorbox{rosegold!40}{their} \colorbox{viridian!30}{music}!\\
    N & ``@\colorbox{rosegold!40}{username}: Karma \colorbox{viridian!60}{has} \colorbox{viridian!30}{no} \colorbox{rosegold!40}{menu}. You get served what you deserve."\\
    \bottomrule
  \end{tabularx}
  \caption{Examples of LIME
    explanations. \colorbox{viridian!30}{Green} indicates a positive
    contribution of the word, and \colorbox{rosegold!40}{red} a
    negative contribution. The augmentation approach used in each
    example is the best-performing method for the respective
    concept. All examples are true positives.}
  \label{tab:lime}
\end{table}

\section{Conclusion \& Future Work}
We outlined a novel methodology for automatic author profiling in
social media users without a costly collection of annotated social
media data. Instead, we directly train on items from validated
psychometric tests. This data selection
procedure has some advantages: items of psychometric tests are
carefully validated textual instances. Such corpora of such
items constitute ``small data'', but are available for a large number
of concepts. Therefore, developing a method to induce classifiers
directly from psychometric tests is also a promising avenue for future
research.

For the tasks of developing models measuring the big
five personality traits, we tested on Twitter data that has been
collected by asking users to fill out a (different) test. The
transfer appears to be achievable, we obtain results for four out of
five personality traits which are \textit{en par} with in-domain
models, using T5 data augmentation (except Openness, which has
very few test instances).

An important remaining research question is how models can be obtained
that show consistently good results across concepts. In a real-world
setup, test data from the target domain would not be available to make
model selection decisions. One way to go might be to combine various
augmentation methods. Another approach would be to use items as prompts
in a zero-shot learning setup.

\section*{Acknowledgements}
This work was supported by Deutsche Forschungsgemeinschaft
(project CEAT, KL 2869/1-2).

\section*{Ethical considerations}
\label{sec:ethical}
The fact that the current research deals with the sensitive topic of
personality warrants for some ethical considerations. First, the
study has been conducted with anonymized publicly available data. We
did not collect data ourselves and importantly the data did not allow to identify subjects. Therefore, it is neither required nor possible to request IRB approval for the current research, given that IRB is concerned with the protection of human subjects. We had no reasons to doubt that the parties, who originally collected the data got IRB approval and informed consent form the participants who provided their data.

However, we acknowledge that automatic systems for personality trait analysis
can be misused. Further, the application of our proposed model
creation strategy can also be used for other more sensible concepts,
for instance regarding mental health. We propose that such systems are
only made available in such a manner that no personalized results can
be retrieved.

\bibliography{lit}
\bibliographystyle{acl_natbib}

\clearpage

\appendix

\onecolumn

\section{Appendix}
\label{sec:appendix}

\subsection{Detailed Results per Class}

\begingroup
\newcommand{\hl}[1]{\textbf{#1}}
\newcommand{\positive}{$+$}
\newcommand{\negative}{$-$}
\newcommand{\bala}[1]{\par{\tiny\scalefont{1.2}\raisebox{1mm}{#1}}}
\newcommand{\neutral}{$\circ$}
\newcommand{\macro}{\textsc{avg}}
\newcommand{\weight}{w-avg}
\newcommand{\ti}[1]{~~\llap{\textendash }~~ #1}
\renewcommand{\ti}[1]{\multicolumn{1}{r}{#1}}
\centering\small
\setlength{\tabcolsep}{3.5pt}
\renewcommand{\arraystretch}{0.9}
\begin{tabular}{ll >{\raggedleft\arraybackslash}p{10mm}>{\raggedleft\arraybackslash}p{10mm}>{\raggedleft\arraybackslash}p{10mm} rrr rrr rrr  rrr   rrr}
  \toprule
  && \multicolumn{12}{c}{\textit{Psychometric Models}} \\
  \cmidrule{3-14}
  && \multicolumn{3}{c}{Plain} &  \multicolumn{3}{c}{EDA}  &  \multicolumn{3}{c}{T5} &  \multicolumn{3}{c}{GPT-2} & \multicolumn{3}{c}{\textit{in-domain}}  & \multicolumn{3}{c}{\textit{Baseline}} \\
  \cmidrule(r){3-5}\cmidrule(lr){6-8}\cmidrule(lr){9-11}\cmidrule(l){12-14}\cmidrule(lr){15-17}\cmidrule{18-20}
  & Class &P&R&\F&P&R&\F&P&R&\F&P&R&\F&P&R&\F&P&R&\F \\
  \multirow{5}{*}{\rt{Open.}} 
  &\ti{\negative}& .01  & 1.00 & .01 & .02  & 1.00 & .04 & .02    & 1.00      & .04      & \hl{.02}       & \hl{1.00}       & \hl{.04}       & .0       & .0       & .0        & .01 & .33 & .03 \\
  &\ti{\positive}& .0  & .0  & .0 & .0 & .0  & .0 & .0    & .0      & .0      & \hl{1.00}      & \hl{.02}      & \hl{.04}      & .99      & 1.00     & .99       & .97 & .50 & .66 \\
  &\ti{\macro}   & .0  & .50  & .01 & .01 & .50  & .02 &.01& .50 & .02 & \hl{.51} & \hl{.51} & \hl{.04} & .49 & .50 & .50  & .49 & .41 & .34 \\
  &\ti{\weight}  & .0  & .01  & .00 & .0  & .02  & .0 &  .0   & .02      & .0      & \hl{.98} & \hl{.04} & \hl{.04} & .97 & .99 & .98  & .95 & .49 & .64 \\
  \cmidrule(r){1-2}\cmidrule(lr){3-5}\cmidrule(lr){6-8}\cmidrule(lr){9-11}\cmidrule(l){12-14}\cmidrule(lr){15-17}\cmidrule{18-20}
  \multirow{5}{*}{\rt{Consc.}} 
  &\ti{\negative} & .07       & .89       & .13       & .08  & 1.00  & .15  & \textbf{.0}   & \textbf{.0}  & \textbf{.0}  & .08   & 1.00   & .14  & 1.00       & .17       & .29        & .05 & .33 & .09 \\
  &\ti{\positive} & .80       & .04     & .07      & 1.00  & .06  & .12  & \textbf{.93}  & \textbf{.99} & \textbf{.96} & 1.00   & .03   & .05  & .96      & 1.00     & .98       & .90 & .48 & .63 \\
  &\ti{\macro}    &  .43 & .46 & .10 & .54 & .53 & .13 & \textbf{.46}  & \textbf{.50} & \textbf{.48} & .54  & .51  & .10 & .96 & .96 & .94  & .47 & .41 & .36 \\
  &\ti{\weight}   &  .75 & .10 & .07 & .93 & .13 & .12 &  \textbf{.86} & \textbf{.92} & \textbf{.89} &  .93 & .10  & .06 & .78 & .80 & .73  & .84 & .47 & .59 \\
  \cmidrule(r){1-2}\cmidrule(lr){3-5}\cmidrule(lr){6-8}\cmidrule(lr){9-11}\cmidrule(l){12-14}\cmidrule(lr){15-17}\cmidrule{18-20}
  \multirow{5}{*}{\rt{Extrav.}} 
  &\ti{\negative} & .09   & 1.00  & .17  & .09 & 1.00 & .17 & \hl{.0}       & \hl{.0}      & \hl{.0}      & .06  & .42  & .11  & 1.00     & .14      & .25      & .08 & .42 & .13 \\
  &\ti{\positive} & .0  & .0 & .0 & .0 & .0 & .0 & \hl{.90}       & \hl{.92}      & \hl{.91}      & .87 & .38 & .53 & .90      & 1.00     & .95      & .89 & .49 & .63 \\
  &\ti{\macro}    &  .05 & .50 & .08 & .05 & .50 & .08 & \hl{.45}  & \hl{.46} & \hl{.45} & .46 & .40 & .32 & .95 & .57 & .60 & .48 & .45 & .38 \\
  &\ti{\weight}   &  .01 & .09 & .02 & .01 & .09 & .02 & \hl{.82}  & \hl{.83} & \hl{.82} & .79 & .38 & .49 & .91 & .91 & .87 & .82 & .48 & .59 \\
  \cmidrule(r){1-2}\cmidrule(lr){3-5}\cmidrule(lr){6-8}\cmidrule(lr){9-11}\cmidrule(l){12-14}\cmidrule(lr){15-17}\cmidrule{18-20}
  \multirow{5}{*}{\rt{Agree.}} 
  &\ti{\negative}  &.10        & 1.00       & .19      & \textbf{.18}  & \textbf{.15}  & \textbf{.17}  & .08  & .31   & .12  & .08  & .62  & .14  & .0     & .0      & .0      & .05 & .23 & .08\\
  &\ti{\positive}  & .0       & .0      & .0      & \textbf{.91} & \textbf{.92} & \textbf{.91} & .88 & .59  & .71 & .77 & .15 & .25 & .89      & 1.00     & .94      & .84 & .46 & .60\\
  &\ti{\macro}     &  .05 & .50 & .09 & \textbf{.54} & \textbf{.54} & \textbf{.54} & .48 & .45  & .42 & .42 & .38 & .19 & .44 & .50 & .47 & .44 & .35 & .34\\
  &\ti{\weight}    & .01 & .10 & .02 & \textbf{.83} & \textbf{.84} & \textbf{.84} & .80 &  .56 & .65 & .70 & .20 & .24 & .79 & .89 & .84 & .76 & .44 & .55\\
  \cmidrule(r){1-2}\cmidrule(lr){3-5}\cmidrule(lr){6-8}\cmidrule(lr){9-11}\cmidrule(l){12-14}\cmidrule(lr){15-17}\cmidrule{18-20}
  \multirow{5}{*}{\rt{Neur.}}
  &\ti{\negative}  & \textbf{.73} & \textbf{.93} & \textbf{.81} & .71 & .78 & .73 & .73 & .90  & .80 & .72      & .91      & .80      & .73      & 1.00      & .84      & .66 & .45 & .53\\
  &\ti{\positive}  & \textbf{.25} & \textbf{.06} & \textbf{.09} & .16 & .10 & .12 & .23  & .08  & .12 & \.10      & .03      & .04       & 1.00      & .12      & .22      & .19 & .36 & .25\\
  &\ti{\macro}     & \textbf{.49} & \textbf{.50} & \textbf{.46} & .43 & .45 & .44 & .47  & .49  & .46 & .41 & .47 & .42  & .86 & .56 & .53 & .43 & .41 & .39\\
  &\ti{\weight}    & \textbf{.60} & \textbf{.70} & \textbf{.63} & .56 & .60 & .58 &  .58 & .68  & .62 & .55 & .66 & .60  & .81      & .74      & .65      & .53 & .43 & .46 \\
  
  \bottomrule
\end{tabular}
\captionof{table}{Detailed results for \textit{Psychometric Models}
  vs. \textit{in-domain Models} vs. \textit{Random Baseline} for
  psychological traits in Twitter users. The random baseline generates
  predictions by respecting the training sets' class distribution. The
  weighted average values for P, R, \F correspond to the average
  across all labels considering the proportion for each label in the
  data set. The bold typo highlights our best performing model
  w.r.t. the highest w-avg. \negative: scored negative, \positive:
  scored positive.}
\label{tab:results}
\endgroup

\subsection{Implementation Details}
We performed the experiments on 4 NVIDIA GeForce GTX 1080 Ti GPUs with
Intel(R) Xeon(R) CPU E5-2650 v4 @ 2.20GH. The number of parameters is
defined by the base model that we used, namely BERT base, with 110 M parameters.

We show the run-time of models (training + testing) in
Table~\ref{tab:runtime}. The numbers do not include startup/loading
times. Note that the test data is (sometimes dramatically) larger than
the training data.\\[\baselineskip]

\begingroup
\centering
\begin{tabular}{lllll}
  \toprule
& \multicolumn{4}{c}{Model} \\
  \cmidrule(l){2-5}
  Concept           & Plain     & EDA      & T5       & GPT-2 \\
  \cmidrule(r){1-1}\cmidrule(rl){2-2}\cmidrule(rl){3-3}\cmidrule(rl){4-4}\cmidrule(l){5-5}
  Depression        & 2460+3900 & 360+3900 & 900+3960 & 900+3960 \\
  Anxiety           & 450+2850  & 407+2905 & 921+2615 & 374+2584 \\
  ADHD              & 550+2650  & 380+2856 & 720+2100 & 875+2175 \\
  \cmidrule(r){1-1}\cmidrule(rl){2-2}\cmidrule(rl){3-3}\cmidrule(rl){4-4}\cmidrule(l){5-5}
  Openness          & 180+120   & 120+120  & 420+120  & 780+120 \\
  Conscientiousness & 480+120   & 180+120  & 780+120  & 360+120 \\
  Extraversion      & 1020+120  & 180+120  & 420+120  & 720+120 \\
  Agreeableness     & 60+120    & 84+120   & 855+120  & 450+120 \\
  Neuroticism       & 474+120   & 85+120   & 400+120  & 650+120 \\
  \bottomrule
\end{tabular}
\captionof{table}{\label{tab:runtime}Runtime of models in seconds (train+test).}
\endgroup

\clearpage

\subsection{Examples for Augmentation Methods}
\label{sec:augmentation}
\begin{itemize}
	\item As an example for the EDA augmentation method, synonym replacement lead to ``Love thinking about
	things'' based on the IPIP-NEO-300 item ``Enjoy thinking
	about things'' for the trait of openness.
	\item As an example for the T5 augmentation method, T5-paraphrasing lead to ``Have fun and be wildly inspired by wild fantasy dreams'' based on the IPIP-NEO-300 item ``Enjoy wild flights of fantasy'' for the trait of openness.
	\item An example for a GPT-2
	generated item measuring agreeableness is ``I am an average person''.
\end{itemize}

\end{document}